\documentclass[runningheads]{llncs}

\usepackage[final,year=2024]{eccv}

\usepackage{eccvabbrv}
\usepackage{multirow}
\usepackage{makecell}
\usepackage{wrapfig}
\usepackage{float}
\usepackage{graphicx}
\usepackage{booktabs}
\usepackage{marvosym}

\usepackage[accsupp]{axessibility}  

\usepackage{hyperref}

\usepackage{orcidlink}

\begin{document}

\title{Momentum Auxiliary Network for Supervised Local Learning} 

\titlerunning{Momentum Auxiliary Network for Supervised Local Learning}

\author{Junhao Su \inst{1}$^*$ \and
Changpeng Cai\inst{1}$^*$ \and
Feiyu Zhu \inst{2}$^*$ \and
Chenghao He \inst{4} \\ 
Xiaojie Xu \inst{5} \and
Dongzhi Guan\textsuperscript{1\Letter} \and
Chenyang Si\textsuperscript{3\Letter}
}

\authorrunning{Junhao Su et al.}

\institute{Southeast University \and
University of Shanghai for Science and Technology \and
Nanyang Technological University \and
East China University of Science and Technology \and
The Hong Kong University of Science and Technology}

\maketitle

\begin{abstract}
\makeatletter{\renewcommand*{\@makefnmark}{}
\footnotetext{$^*$Equal Contribution. \\
\Letter Corresponding Authors: Chengyang Si (chenyang.si.mail@gmail.com) and Dongzhi Guan (guandongzhi@seu.edu.cn).}}

Deep neural networks conventionally employ end-to-end backpropagation for their training process, which lacks biological credibility and triggers a locking dilemma during network parameter updates, leading to significant GPU memory use. Supervised local learning, which segments the network into multiple local blocks updated by independent auxiliary networks. However, these methods cannot replace end-to-end training due to lower accuracy, as gradients only propagate within their local block, creating a lack of information exchange between blocks.
To address this issue and establish information transfer across blocks, we propose a Momentum Auxiliary Network (MAN) that establishes a dynamic interaction mechanism. The MAN leverages an exponential moving average (EMA) of the parameters from adjacent local blocks to enhance information flow. This auxiliary network, updated through EMA, helps bridge the informational gap between blocks. Nevertheless, we observe that directly applying EMA parameters has certain limitations due to feature discrepancies among local blocks. To overcome this, we introduce learnable biases, further boosting performance. We have validated our method on four image classification datasets (CIFAR-10, STL-10, SVHN, ImageNet), attaining superior performance and substantial memory savings. Notably, our method can reduce GPU memory usage by more than 45\% on the ImageNet dataset compared to end-to-end training, while achieving higher performance. The Momentum Auxiliary Network thus offers a new perspective for supervised local learning. Our code is available at: \url{https://github.com/JunhaoSu0/MAN}.

\end{abstract}
\section{Introduction}
\label{sec:intro}

In deep learning, the prevalent use of end-to-end backpropagation \cite{2} is essential for training complex neural networks, which requires heavy computational work for loss evaluation and successive gradient descent through network layers to refine parameters \cite{20, 21, 14}. This method starkly contrasts with the local signal processing of biological synapses \cite{7,35,36} and imposes a 'locking' constraint \cite{1} that delays parameter updates until the entire forward and backward pass is complete. Such constraints exacerbate challenges like reduced parallelism and increased GPU memory usage \cite{16,18}, compromising training efficiency and scalability.
\begin{figure}[t]
    \centering
    \includegraphics[width=\textwidth]{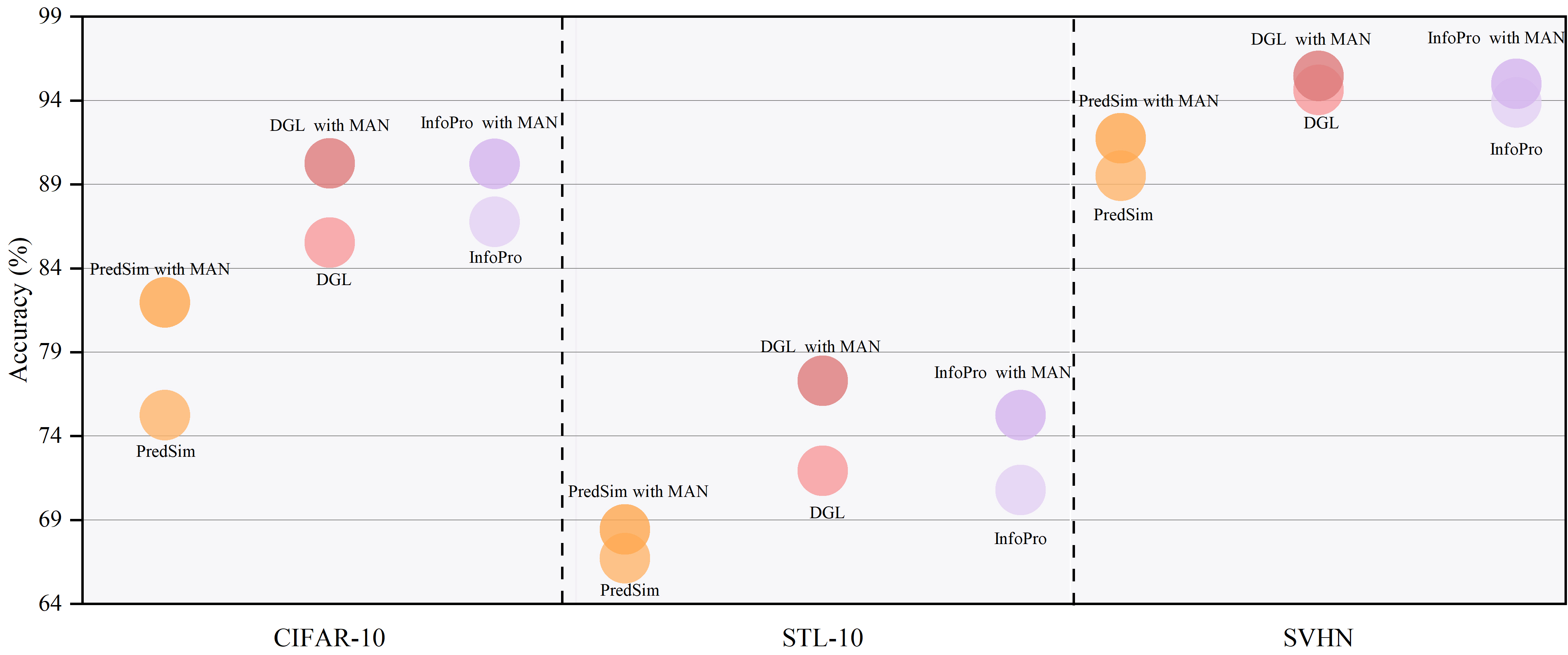}
    \caption{Comparison between different methods with MAN and the original methods in terms of accuracy. Results are obtained using ResNet-110 (K=55) on the various datasets.}
    \label{Figure 1}
\end{figure}
An emerging alternative to traditional end-to-end (E2E) training is Local Learning \cite{1,11,12,13,14,15,27,28,43}, which promises to mitigate the drawbacks of traditional methods. These methods divide the network into discrete, gradient-isolated blocks, each of which is updated by its dedicated auxiliary network according to distinct local objectives \cite{15,27,28}. This training strategy enables immediate parameter updates upon local error reception, circumventing the sequential update bottleneck of E2E training, thereby significantly enhancing the efficiency of parallel training \cite{45,46}. Moreover, local learning reduces the demand on GPU memory by only retaining gradients for the local and auxiliary networks, thus eliminating the need to store extensive global gradient information and saving computational resources \cite{16,18}. However, even with these advantages, there is still a significant performance gap between local learning and traditional E2E methods, especially as the network is divided into more local blocks. Current local learning techniques mainly focus on improving the design of auxiliary network structures \cite{28} and making local loss functions better to close this performance gap\cite{15,27}. Yet, these improvements do not completely address the inherent short-sightedness of local learning: the separation into blocks can make each part of the network only focus on its local objectives, possibly ignoring the overall objectives of the network. This can lead to the discarding of globally beneficial information due to the lack of inter-block communication.

In this paper, we introduce the Momentum Auxiliary Network (MAN), a novel network architecture designed to mitigate the inherent limitations of supervised local learning by enhancing inter-block communication. 
Specifically, MAN not only accepts the current local block as input but also absorbs the parameters of its next block. Upon completion of a forward pass, parameter updates are performed using local gradients refined through Exponential Moving Average (EMA) techniques \cite{34}. This novel approach enables each local block to integrate information from the following block, thereby extending the operational perspective of each block beyond its immediate objectives and further aligning with the global objective of the network. This addresses the critical limitations in local learning frameworks, ensuring a more cohesive alignment with the overarching network objectives.
Furthermore, we identify that directly applying EMA parameters is constrained by the feature incongruities across gradient-isolated blocks.
To mitigate this, we implement additional learnable bias terms to each auxiliary block, enhancing their ability to share information effectively. 
The proposed MAN requires only a minimal increase in memory use but significantly improves performance by enhancing the information sharing ability of local blocks 
The efficacy of the MAN approach is validated on a suite of benchmark image classification datasets, 
including CIFAR-10 \cite{25}, STL-10 \cite{31}, SVHN \cite{30}, and ImageNet \cite{32}.
The results demonstrate the effectiveness of our method in surpassing the limitations of traditional supervised local learning.

The contributions of this paper could be summarized as follows:
\begin{itemize}
\item We propose Momentum Auxiliary Network (MAN), designed to facilitate inter-block communication with Exponential Moving Average (EMA), mitigating the short-sightedness issue in conventional supervised local learning techniques and culminating in an elevated overall performance of the network.

\item MAN is a versatile, plug-and-play approach that can seamlessly integrate with any supervised local learning technique, markedly amplifying their efficacy while requiring only a negligible increase in memory usage.

\item MAN has demonstrated its effectiveness through experiments on benchmark image classification datasets, achieving state-of-the-art performance. Notably, on the ImageNet \cite{32} dataset, it outperforms E2E training while using significantly less memory.

\end{itemize}
\begin{figure*}[t]
    \centering
    \includegraphics[width=\textwidth]{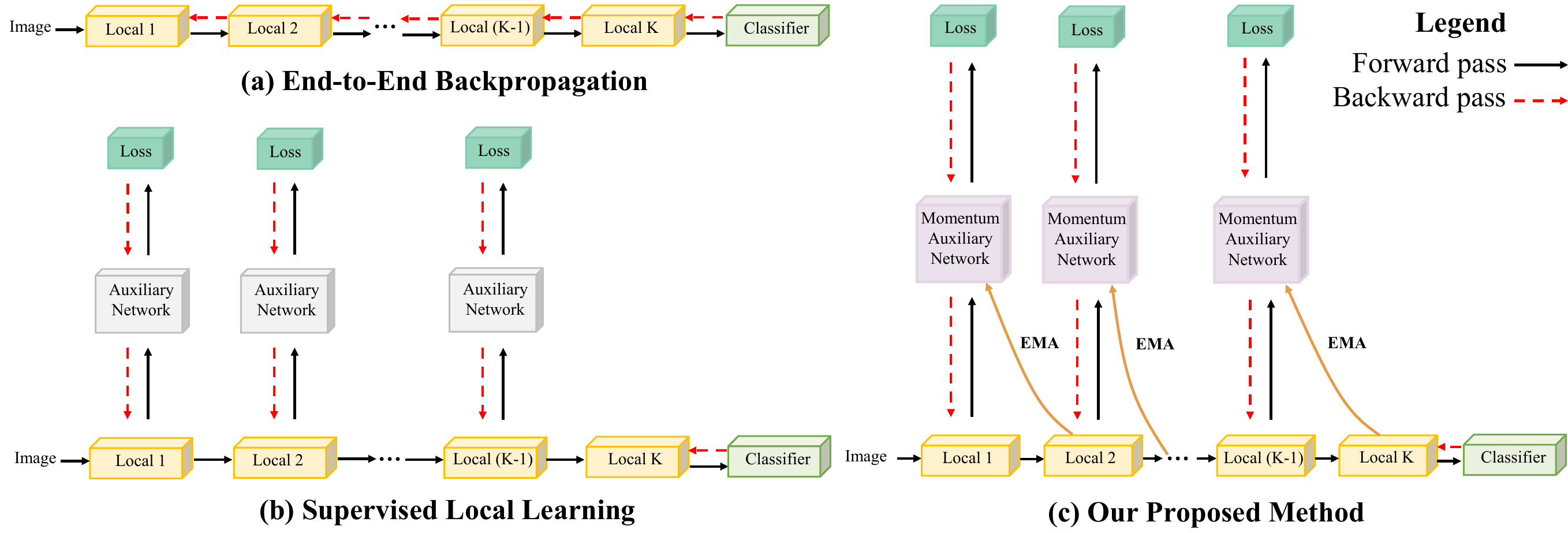}
    \caption{Comparison of (a) end-to-end backpropagation, (b) other supervised local learning methods, and (c) our proposed method. Unlike E2E, supervised local learning separates the network into K gradient-isolated local blocks.}
    \label{Figure 2}
\end{figure*}
 
\section{Related Work}
\label{sec:formatting}
\subsection{Local Learning}
Local learning is first proposed as an innovative deep learning algorithm with the intention of utilizing memory more efficiently and adhering more closely to principles of biological plausibility \cite{9}. This approach emerges as a response to the limitations presented by global E2E training \cite{41}, prompting the development of alternative supervised local learning methodologies. Examples of these methodologies include a differentiable search algorithm, which decouples network blocks for block-level learning and selects a manually assisted network for each local block \cite{42}, as well as a self-supervised contrasting loss function that leverages its local learning rules \cite{13,43}. In supervised local learning, training primarily relies on the design of reasonable supervised local loss functions or the manual construction of auxiliary networks.

\noindent {\bfseries Supervised Local Loss:} InfoPro \cite{27} presents a method that involves the reconstruction loss for each hidden layer, coupled with cross-entropy loss. This approach forces the hidden layers to preserve vital input information, thereby preventing the early layers from discarding information relevant to the task. PredSim \cite{15} employs layer-wise loss functions for network training. It makes use of two distinct supervised loss functions to generate local error signals for hidden layers. The optimization of these losses takes place during supervised local learning, thus eliminating the necessity to propagate global errors back to the hidden layers. These methods have achieved advanced performance in the field of supervised local learning, but there is still a significant gap compared to the performance of E2E training.

\noindent {\bfseries Auxiliary Networks:} InfoPro \cite{27} incorporates two auxiliary networks into its approach. One network consists of a single convolutional layer followed by two fully-connected layers, which are employed for the cross-entropy loss. The other network utilizes two convolutional layers with upsampling for the reconstruction loss. Alternatively, DGL \cite{28} designs a structure that uses three consecutive convolutional layers connected to a global pooling layer, followed by three consecutive fully connected layers. Moreover, DGL restricts the parameter of the auxiliary network to only 5\% of the main network capacity, offering advantages in terms of speed and memory savings. However, these method still perform poorly when the network is divided into a large number of local blocks.

\subsection{Alternative Learning Rules to E2E Training}
Due to certain limitations inherent in E2E training, the pursuit of alternative methods to E2E training has gradually garnered attention in recent years \cite{10}. Several efforts have been dedicated to addressing the biologically implausible aspects of E2E training, such as the training methodology of target propagation \cite{51,52,53} attempts to train a specialized backward connection by utilizing local reconstruction targets. Additionally, some recent studies have attempted to completely avoid backpropagation in neural networks through forward gradient learning \cite{54,55}. Meanwhile, the weight transport problem \cite{9}. This has been tackled either by employing distinct feedback connections \cite{47,48} or by directly disseminating global errors to each hidden unit \cite{49,50}. While these methods have somewhat alleviated the biological implausibility of E2E training, they still rely on global objectives, which fundamentally differ from biological neural networks that rely on local synapses for information transmission. Furthermore, these methods currently struggle to be effective on large datasets \cite{32}.

\section{Method}
\subsection{Preliminaries}

To set the stage, we first provide a brief overview of traditional end-to-end supervised learning and backpropagation mechanisms. We denote a data sample as $x$ and its corresponding ground-truth label as $y$. The entire deep network is fragmented into several local blocks. During the forward propagation process, the output from the j-th block serves as the input for the (j+1)-th block, expressed as $x_{j+1}=f_{\theta_j}(x_j)$. Here, $\theta_{j}$ symbolizes the parameters of the j-th local block and $f(\cdot)$ represents the forward calculation of the block. We evaluate the loss function ${\mathcal{L}}(\hat{y}, y)$ between the output of the last block and the ground truth label, and propagate it back iteratively to the preceding blocks.

Supervised local learning strategies \cite{15,27,28} integrate auxiliary networks for local supervision. For each local block, an auxiliary network is affixed. The output from a local block is fed to its corresponding auxiliary network, generating the local supervisory signal as $\hat{y_{j}}=g_{\gamma_j}(x_{j+1})$. Here, $\gamma_{j}$ denotes the parameters of the j-th auxiliary network.

In this setup, we update the parameters of the j-th auxiliary network and local block, $\gamma_j, \theta_j$, as follows:

\begin{equation}
\gamma_j \leftarrow \gamma_j - \eta_a \times \nabla_{\gamma_j} \mathcal{L}(\hat{y_j}, y)
\end{equation}

\begin{equation}
\theta_j \leftarrow \theta_j - \eta_l \times \nabla_{\theta_j} \mathcal{L}(\hat{y_j}, y)
\end{equation}

\noindent where $\eta_a, \eta_l$ are the learning rates of the auxiliary networks and local blocks, respectively. By attaching auxiliary networks, each local block becomes gradient-isolated and can be updated with local supervision rather than global backpropagation.

\subsection{Momentum Auxiliary Network}

Existing techniques incorporate supervision signals into individual local blocks, which allows for parallel parameter updates and reduces memory overhead. However, this approach can lead to a short-sightedness problem, where each local block neglects the information from subsequent blocks, ultimately resulting in suboptimal final accuracy.

To address this issue, we propose a comprehensive information exchange module—the Momentum Auxiliary Network (MAN). The MAN employs the Exponential Moving Average (EMA) mechanism \cite{34} as a conduit for transferring information from subsequent blocks to the current block. In the context of MAN, we update the parameters of the j-th auxiliary network and local block as follows:

\begin{equation}
\gamma_j \leftarrow \gamma_j - \eta_a \times \nabla_{\gamma_j} \mathcal{L}(\hat{y_j}, y)
\end{equation}

\begin{equation}
\gamma_j \leftarrow EMA(\gamma_j, \theta_{j+1})
\end{equation}

\begin{equation}
\theta_j \leftarrow \theta_j - \eta_l \times \nabla_{\theta_j} \mathcal{L}(\hat{y_j}, y)
\end{equation}

\noindent where $\gamma_j$ represents the parameters of the j-th auxiliary network and $\theta_j$ represents those of the j-th local block. After updating with local gradients, $\gamma_j$ undergoes further refinement by incorporating the parameters of the subsequent local block via the EMA, which is a weighted sum operation. 

\begin{wrapfigure}[15]{r}{0.5\textwidth}
    \centering
    \includegraphics[width=\linewidth]{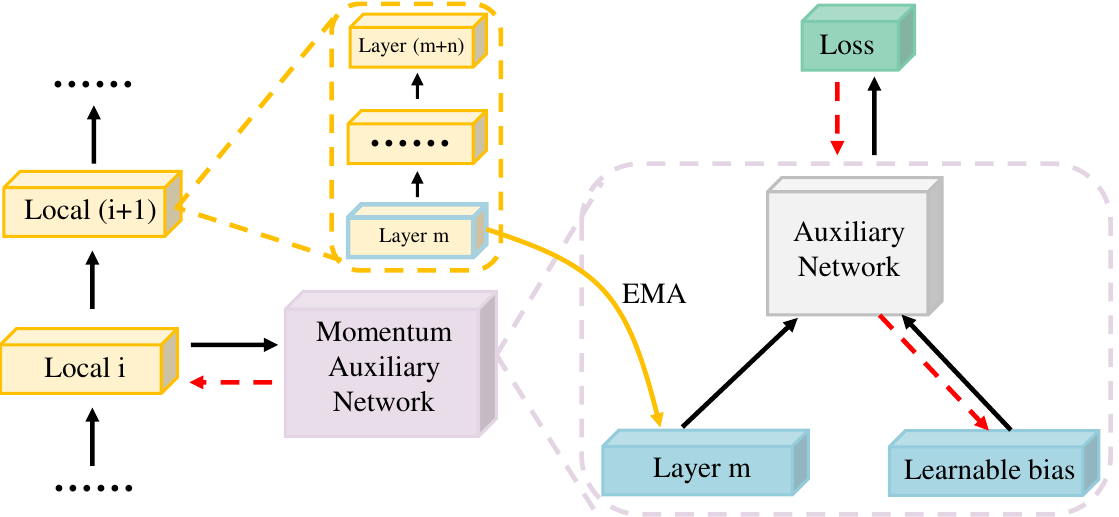}
    \caption{Details of the Momentum Auxiliary Network. Local (i+1) represents the (i+1)-th gradient-isolated local block, which contains layers from layer m to layer (m+n), totaling n+1 layers (n$\geqslant$0). We only use the parameters of the first layer to ensure a balance in GPU memory usage.}
    \label{Figure 3}
\end{wrapfigure} 

However, through experimental results, it becomes apparent that directly applying EMA parameters to update the auxiliary network has its limitations, and it provides limited improvements to each local block. Upon analyzing, we find that the features learned between each local block vary to a certain degree, which hampers the effectiveness of the EMA update method. As a consequence, a learnable bias is introduced to augment the learning capability of the hidden layers within the local block, while also compensating for the deficiencies of the EMA update method. \\ Therefore, the parameter update method for the j-th auxiliary network can be written as:

\begin{equation}
(\gamma_j,b_j) \leftarrow (\gamma_j,b_j) - \eta_a \times \nabla_{(\gamma_j,b_j)} \mathcal{L}(\hat{y_j}, y)
\end{equation}

\begin{equation}
\gamma_j \leftarrow EMA(\gamma_j, \theta_{j+1})
\end{equation}

\noindent where ($\gamma_j,b_j$) represent the parameters of the j-th auxiliary network. During this process, we update the parameters $\gamma_j$ of the j-th auxiliary network jointly through the learnable bias $b_j$ and EMA.

Essentially, the proposed MAN facilitates communication between subsequent blocks through the EMA mechanism \cite{34}, effectively resolving the short-sightedness problem inherent in traditional supervised local learning methods. Concurrently, it introduces an independent learnable bias to mitigate information discrepancies caused by feature variations among different local blocks, thereby aligning the output of each block closer to the global target.

In addition to the two innovative methods mentioned above, MAN demonstrates strong versatility. It can be seamlessly integrated into existing supervised local learning methods, and it excels across various datasets, reflecting its flexibility and generalizability.

\section{Experiments}

\subsection{Experimental Setup}
We conduct experiments using four widely adopted datasets: CIFAR-10 \cite{25}, SVHN \cite{30}, STL-10 \cite{31}, and ImageNet \cite{32}, with ResNets \cite{24} of varying depths serving as the network architectures.

Three state-of-the-art supervised local learning methods are selected for comparison: PredSim \cite{15}, DGL \cite{28}, and InfoPro \cite{27}. We then partition the network into K local blocks, each containing an approximately equal number of layers. Our proposed Momentum Auxiliary Network is incorporated only into the first K-1 local blocks. The K-th local block does not employ an auxiliary network and is directly connected to the output classifier. We compare these configurations against traditional E2E and original supervised local learning methods to maintain consistent training settings and eliminate confounding variables.

\subsection{Implement Details}
\label{detail}
In our experiments on CIFAR-10 \cite{25}, SVHN \cite{30}, and STL-10 \cite{31} datasets with ResNet-32\cite{24} and ResNet-110 \cite{24}, we utilize the SGD optimizer with Nesterov momentum set at 0.9 and an L2 weight decay factor of 1e-4. We employ batch sizes of 1024 for CIFAR-10 and SVHN and 128 for STL-10. The training duration spans 400 epochs, starting with initial learning rates of 0.8 for CIFAR-10 / SVHN and 0.1 for STL-10, following a cosine annealing scheduler \cite{31}.

In our experiments with ImageNet \cite{32}, we adopt different training settings for various architectures. We train VGG13 \cite{33} for 90 epochs with an initial learning rate of 0.025. For ResNet-101 \cite{24}, ResNet-152 \cite{24}, and ResNeXt-101,32 $\times$ 8d \cite{xie2017aggregated}, we train them for 90 epochs as well, with initial learning rates of 0.05, 0.05, and 0.025, respectively. We set the batch sizes to 64 for VGG13, 128 for ResNet-101 and ResNet-152, and 64 for ResNeXt-101,32 $\times$ 8d. We maintain consistency with other training configurations as previously described for CIFAR-10 \cite{25}.

Based on our multiple experimental results, we select the hyperparameter $momentum$ value as 0.995, due to its consistently stable and superior performance. We will provide the experimental results of other $momentum$ hyperparameters in the supplementary materials.

\subsection{Results on Image Classification Datasets}

\noindent \textbf{Results on Image Classification Benchmarks:} We start by assessing the accuracy performance of our approach using the CIFAR-10 \cite{25}, SVHN \cite{30}, and STL-10 \cite{31} datasets. We employ ResNet-32 \cite{24}, partitioned into 8 and 16 local blocks, and ResNet-110 \cite{24}, divided into 32 and 55 local blocks. As illustrated in Table \ref{Table 1}, our MAN significantly bolsters the accuracy of all methods.
\begin{table}[H]
	\centering
\caption{Perfomance of different networks with varying numbers of local blocks. The average test error is obtained by 5 experiments. The {\bfseries *} means addition of our MAN.}
\resizebox{\textwidth}{!}{
	\begin{tabular}{c|ccccc}\hline
\multirow{2}{*}{Dataset}&\multirow{2}{*}{Method}&\multicolumn{2}{c}{ResNet-32}&\multicolumn{2}{c}{ResNet-110}\\\cline{3-6}
&&K = 8 (Test Error)&K = 16 (Test Error)&K = 32 (Test Error)&K = 55 (Test Error)\\ \hline
\multirow{6}{*}{\makecell[c]{CIFAR-10\\ (E2E(ResNet-32)=6.37,\\ E2E(ResNet-110)=5.42)}}&PredSim\cite{15}&20.62&22.71&22.08&24.74\\
&\textbf{PredSim*}&\textbf{14.29($\downarrow$6.33)}&\textbf{15.58($\downarrow$7.13)}&\textbf{17.05($\downarrow$5.03)}&\textbf{18.02($\downarrow$6.72)}\\
&DGL\cite{28}&11.63&14.08&12.51&14.45\\
&\textbf{DGL*}&\textbf{8.42($\downarrow$3.21)}&\textbf{9.11($\downarrow$4.97)}&\textbf{9.65($\downarrow$2.86)}&\textbf{9.73($\downarrow$4.72)}\\
&InfoPro\cite{27}&11.51&12.93&12.26&13.22\\
&\textbf{InfoPro*}&\textbf{9.32($\downarrow$2.19)}&\textbf{9.65($\downarrow$3.28)}&\textbf{9.06($\downarrow$3.20)}&\textbf{9.77($\downarrow$3.45)}\\\hline
\multirow{6}{*}{\makecell[c]{STL-10 \\(E2E(ResNet-32)=19.35,\\ E2E(ResNet-110)=19.67)}}&PredSim\cite{15}&31.97&32.90&32.05&33.27\\
&\textbf{PredSim*}&\textbf{29.97($\downarrow$2.00)}&\textbf{29.99($\downarrow$2.91)}&\textbf{30.48($\downarrow$1.57)}&\textbf{31.55($\downarrow$1.72)}\\
&DGL\cite{28}&25.05&27.14&25.67&28.16\\
&\textbf{DGL*}&\textbf{20.74($\downarrow$4.31)}&\textbf{21.37($\downarrow$5.77)}&\textbf{22.54($\downarrow$3.13)}&\textbf{22.69($\downarrow$5.47)}\\
&InfoPro\cite{27}&27.32&29.28&28.58&29.20\\
&\textbf{InfoPro*}&\textbf{23.17($\downarrow$4.15)}&\textbf{23.54($\downarrow$5.74)}&\textbf{24.08($\downarrow$4.50)}&\textbf{24.74($\downarrow$4.46)}\\\hline
\multirow{6}{*}{\makecell[c]{SVHN \\(E2E(ResNet-32)=2.99,\\ E2E(ResNet-110)=2.92)}}&PredSim\cite{15}&6.91&8.08&9.12&10.47\\
&\textbf{PredSim*}&\textbf{5.54($\downarrow$1.37)}&\textbf{6.39($\downarrow$1.69)}&\textbf{7.27($\downarrow$1.85)}&\textbf{8.24($\downarrow$2.23)}\\
&DGL\cite{28}&4.83&5.05&5.12&5.36\\
&\textbf{DGL*}&\textbf{3.80($\downarrow$1.03)}&\textbf{4.04($\downarrow$1.01)}&\textbf{4.08($\downarrow$1.04)}&\textbf{4.52($\downarrow$0.84)}\\
&InfoPro\cite{27}&5.61&5.97&5.89&6.11\\
&\textbf{InfoPro*}&\textbf{4.49($\downarrow$1.12)}&\textbf{5.19($\downarrow$0.78)}&\textbf{4.85($\downarrow$1.04)}&\textbf{4.99($\downarrow$1.12)}\\\hline
	\end{tabular}}
    \label{Table 1}
\end{table}

On the CIFAR-10 dataset \cite{25}, our method exhibits considerable improvements in diminishing test errors across various methods. In the relatively shallower network of ResNet-32 (K=16), where individual layers function as gradient-isolated local blocks, we record a reduction in test errors for PredSim \cite{15}, DGL \cite{28}, and InfoPro \cite{27}, from 22.71, 14.08, and 12.93 to 15.58, 9.11, and 9.65 respectively. This translates to a performance enhancement exceeding 25\% for all methods. Even though the performance across all methods in the comparatively deeper network, ResNet-110 (K=55), is somewhat inferior due to the inherent need for more global information in such networks, our method still delivers exceptional performance. It achieves approximately a 20\% improvement across all methods, underscoring the robust effectiveness of MAN in deeper networks.

When applied to other datasets, MAN can also reduce the test error of PredSim \cite{15}, DGL \cite{28}, and InfoPro \cite{27} by at least 5\%, 12\%, and 16\% on the STL-10 \cite{31} dataset. On the SVHN \cite{30} dataset, our improvements over the three methods also surpass 20\%, 21\%, and 13\%. As can be seen, the improvement our MAN introduces to all methods is quite remarkable—comparable even to the accuracy of E2E training—and it significantly mitigates the underwhelming performance issue that has continually plagued supervised local learning.

\noindent \textbf{Results on ImageNet:} We further validate the effectiveness of our approach on ImageNet \cite{32} using four networks of varying depths (ResNets \cite{24} and VGG13 \cite{33}). As depicted in Table \ref{Table 2}, when we employ VGG13 as the backbone and divide the network into 10 blocks, DGL \cite{28} achieves merely a Top1-Error of 35.60 and a Top5-Error of 14.2, representing a substantial gap when compared to the E2E method. However, with the introduction of our MAN, the Top1-Error reduces by 3.61 points, and the Top5-Error decreases by 3.36 points for DGL. This significant enhancement brings the performance closer to the E2E method.

As illustrated in Table \ref{Table 2}, when we use ResNet-101 \cite{24}, ResNet-152 \cite{24}, ResNeXt-101, 32$\times$8d \cite{xie2017aggregated} as backbones and divide the network into four blocks, the performance of InfoPro \cite{27} is already below that of E2E. After incorporating our MAN, the Top-1 Error of these three backbone networks can be reduced by approximately 6\% compared to the original, surpassing the performance of E2E training. These results underscore the effectiveness of our MAN on the large-scale ImageNet \cite{32} dataset, even when using deeper networks.
\begin{table}[H]
\tabcolsep 0.5cm
\scriptsize
	\centering
\caption{Results on the validation set of ImageNet}

\begin{tabular}{cccc}\hline
Network&Method&Top1-Error&Top5-Error\\ \hline
\multirow{5}{*}{\makecell[c]{ResNet-101}}  &E2E&22.03&5.93\\
&InfoPro(K=2)\cite{27}&21.85&5.89\\
&\textbf{InfoPro*(K=2)}&\textbf{21.65($\downarrow$0.20)}&\textbf{5.49($\downarrow$0.40)}\\
&InfoPro(K=4)\cite{27}	&22.81	&6.54\\
&\textbf{InfoPro*(K=4)}	&\textbf{21.73(↓1.08)}	&\textbf{5.81(↓0.73)}\\
\hline
\multirow{5}{*}{\makecell[c]{ResNet-152}}  &E2E&21.60&5.92\\
&InfoPro(K=2)\cite{27}&21.45&5.84\\
&\textbf{InfoPro*(K=2)}&\textbf{21.23($\downarrow$0.22)}&\textbf{5.53($\downarrow$0.31)}\\
&InfoPro(K=4)\cite{27}	&22.93&	6.71\\
&\textbf{InfoPro*(K=4)}	&\textbf{21.59(↓1.34)}	&\textbf{5.89(↓0.82)}\\
\hline
\multirow{5}{*}{\makecell[c]{ResNeXt-101,\\ 32 $\times$ 8d}}  &E2E&20.64&5.40\\
&InfoPro(K=2)\cite{27}&20.35&5.28\\
&\textbf{InfoPro*(K=2)}&\textbf{20.11($\downarrow$0.24)}&\textbf{5.18($\downarrow$0.10)}\\
&InfoPro(K=4)\cite{27}	&21.69	&6.11\\
&\textbf{InfoPro*(K=4)}	&\textbf{20.37(↓1.32)}&	\textbf{5.34(↓0.77)}\\
\hline
\multirow{3}{*}{\makecell[c]{VGG13}}&E2E&28.41&9.63\\
&DGL\cite{28}&35.60&14.20\\
&\textbf{DGL*(K=10)}&\textbf{31.99($\downarrow$3.61)}&\textbf{10.84($\downarrow$3.36)}\\\hline  
\end{tabular} 
    \label{Table 2} 
\end{table}

\begin{figure}[htbp]
    \centering
    \includegraphics[width=\textwidth]{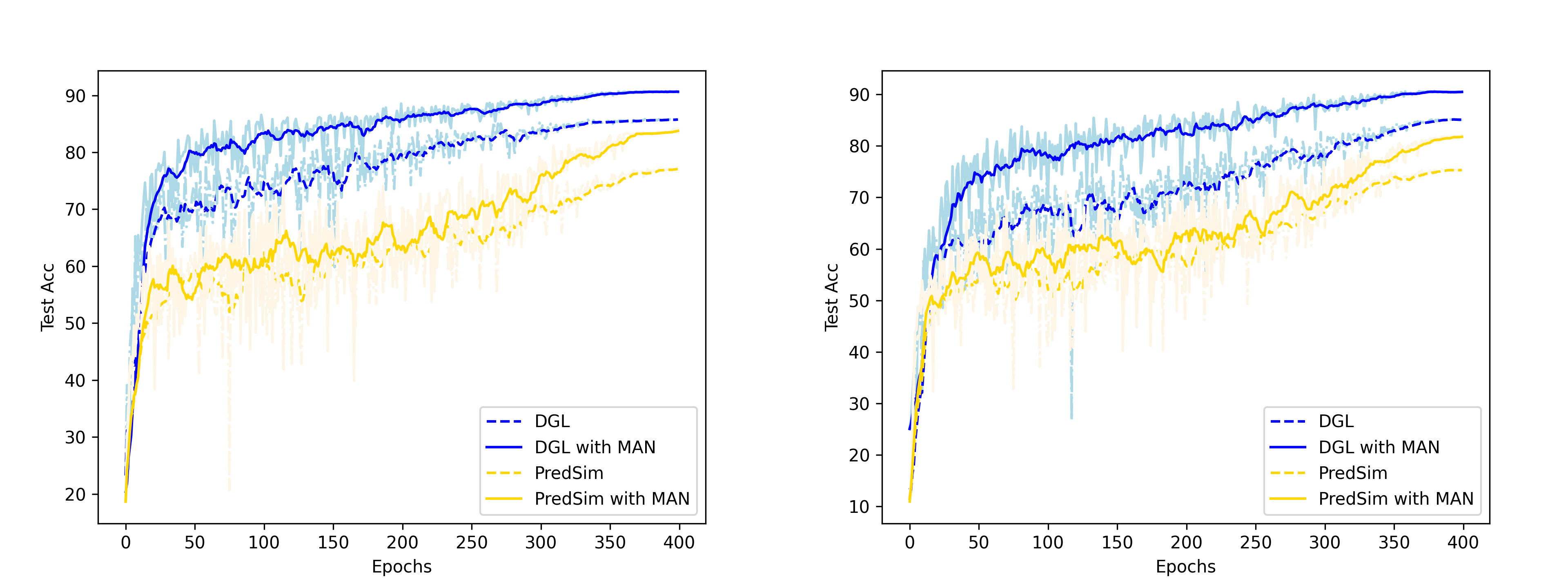}
    \caption{Training-Accuracy curves, the left uses ResNet-32 (K=16) as the backbone, while the right uses ResNet-110 (K=55). Both are utilizing the CIFAR-10 dataset.}
    \label{Figure 4}
\end{figure}

\noindent \textbf{Training-Accuracy Curve Analysis:} As depicted in the accuracy-epoch curve in Fig. \ref{Figure 4}, our Momentum Auxiliary Network consistently outperforms the original method in terms of accuracy throughout the entire training process. This underscores its reliability and stability during the training process of classification tasks. Moreover, our MAN achieves a higher accuracy earlier in the later stages of training and attains stability sooner, indicating a faster convergence rate—a critical attribute in large-scale and complex tasks.

In summary, our Momentum Auxiliary Network significantly enhances the accuracy of traditional supervised local learning methods by promoting information exchange between local blocks. 
The beneficial information exchange facilitated by Momentum Auxiliary Network significantly mitigates the pervasive issue of shortsightedness within the supervised local learning domain, thereby offering advantages in terms of training stability and convergence speed. This is crucial for training models more effectively and efficiently.

\begin{table}[H]
    \centering 
    \caption{Comparison of GPU memory usage using InfoPro as baseline with different backbones on the ImageNet dataset.}
    \scalebox{0.95}{{}\begin{tabular}{cccc}\toprule
Method & \begin{tabular}{c} 
ResNet-101 \\
GPU Memory(GB)
\end{tabular} & \begin{tabular}{c} 
ResNet-152 \\
GPU Memory(GB)
\end{tabular} & \begin{tabular}{c} 
ResNeXt-101, 32×8d \\
GPU Memory(GB)
\end{tabular} \\ \midrule
E2E & 19.71 & 26.29 & 19.22 \\
\midrule InfoPro(K=2)\cite{27} & 12.06($\downarrow$ 38.8\%) & 15.53($\downarrow$ 40.9\%) & 11.55($\downarrow$ 39.9\%) \\
\bf InfoPro*(K=2) & $\mathbf{1 2 . 3 2(\downarrow 37.5\%)}$ & $\mathbf{1 5 . 9 3(\downarrow 39.4\%)}$ & $\mathbf{1 1 . 7 3(\downarrow 38.9\%)}$ \\
\midrule InfoPro(K=4)\cite{27} & 10.37($\downarrow$ 47.3\%) & 13.48($\downarrow$ 48.7\%) & 10.24($\downarrow$ 46.7\%) \\
\bf InfoPro*(K=4) & $\mathbf{1 0 . 6 9(\downarrow 45.8\%)}$ & $\mathbf{1 3 . 9 1(\downarrow 47.1\%)}$ & $\mathbf{1 0 . 4 9(\downarrow 45.5\%)}$ \\
\bottomrule
\end{tabular}}
    \label{Table 3}
\end{table}

\noindent \textbf{Results on GPU memory requirement:} Supervised local learning markedly conserves GPU memory by limiting gradient propagation to within local blocks, thereby ensuring that backpropagation transpires exclusively within each respective local block and its corresponding auxiliary network. This strategy significantly diminishes the storage demands for activation parameters and gradient information, which would typically disseminate across local blocks. Consequently, this facilitates substantial savings in GPU memory utilization. Our analysis, featuring a comprehensive comparison of GPU memory usage on the ImageNet dataset, illustrates that the implementation of a Momentum Auxiliary Network can elicit notable performance enhancements while incurring only a minimal uptick in GPU memory consumption.

As shown in Table \ref{Table 3}, it indicates the substantial GPU memory reduction on the ImageNet\cite{32}. When applying our method to InfoPro \cite{27} on ResNet-101 \cite{24}, ResNet-152 \cite{24}, and ResNeXt-101, 32$\times$8d \cite{xie2017aggregated}, dividing the backbone into two local blocks results in a GPU memory saving of 37.5\% to 39.4\%. When the network is divided into four local blocks, the degree of GPU memory savings is further enhanced, exceeding 45\%. Combined with Table \ref{Table 2}, it can be seen that we achieve better performance with almost half the GPU memory required for end-to-end training. Further analysis of the memory usage comparison with the original method shows that our approach only increases the GPU memory by about 1\% over the original method, while achieving a performance improvement of over 5\%. These results highlight the excellent balance that the MAN method achieves in terms of GPU memory usage and performance enhancement.

\subsection{Ablation Studies}
 \begin{table}[t]
	\centering
 \caption{Abalation study of MAN. (a) Using DGL as baseline and ResNet-32 (K=16) as backbone on the CIFAR-10 dataset. (b) Using InfoPro as baseline and ResNet-101 (K=4) as backbone on the ImageNet dataset. LB stands for Learnable Bias.}
\begin{minipage}{0.49\linewidth}
\begin{tabular}{ccc} 
\hline EMA & LB & Test Error \\
\hline $\times$ & $\times$ & 14. 08 \\
$\checkmark$ & $\times$ & $\boldsymbol{11.07(\downarrow 3.01)}$ \\
$\checkmark$ & $\checkmark$ & $\boldsymbol{9.11(\downarrow 4.97)}$ \\
\hline
\multicolumn{3}{c}{(a)}
\end{tabular}
\end{minipage}
\begin{minipage}{0.49\linewidth}
\begin{tabular}{cccc} 
\hline EMA & LB & Top1-Error & Top5-Error \\
\hline $\times$ & $\times$ & 22.81 & 6.54 \\
$\checkmark$ & $\times$ & $\boldsymbol{22.09(\downarrow 0.72)}$ & $\boldsymbol{6.07(\downarrow 0.47)}$ \\
$\checkmark$ & $\checkmark$ & $\boldsymbol{21.73(\downarrow 1.08)}$ & $\boldsymbol{5.81(\downarrow 0.73)}$ \\
\hline
\multicolumn{4}{c}{(b)}
\end{tabular}
\end{minipage}
\label{Table 4}
\end{table}

We conduct an ablation study on the CIFAR-10 \cite{25} dataset and ImageNet dataset to assess the impact of the EMA method \cite{34} and learnable bias in the MAN on performance. For this analysis, we use ResNet-32 (K=16) \cite{24} as the backbone and the original DGL \cite{28} method as a comparison baseline.

As shown in Table \ref{Table 4}(a), when we only use the EMA \cite{34} in MAN to promote information exchange with subsequent blocks, without adding a learnable bias, the test error decreases from 14.08 to 11.07. When we further add a learnable bias, the test error decreases from 11.07 to 9.11, it is evident that both the EMA method and learnable bias contribute to performance improvement. We hypothesize that the EMA and learnable bias are somewhat complementary in terms of the features they learn. To validate this, we conduct feature visualization.

\begin{figure}[H]
\centering
	\begin{minipage}{0.24\linewidth}
		\centering
		\includegraphics[width=\linewidth]{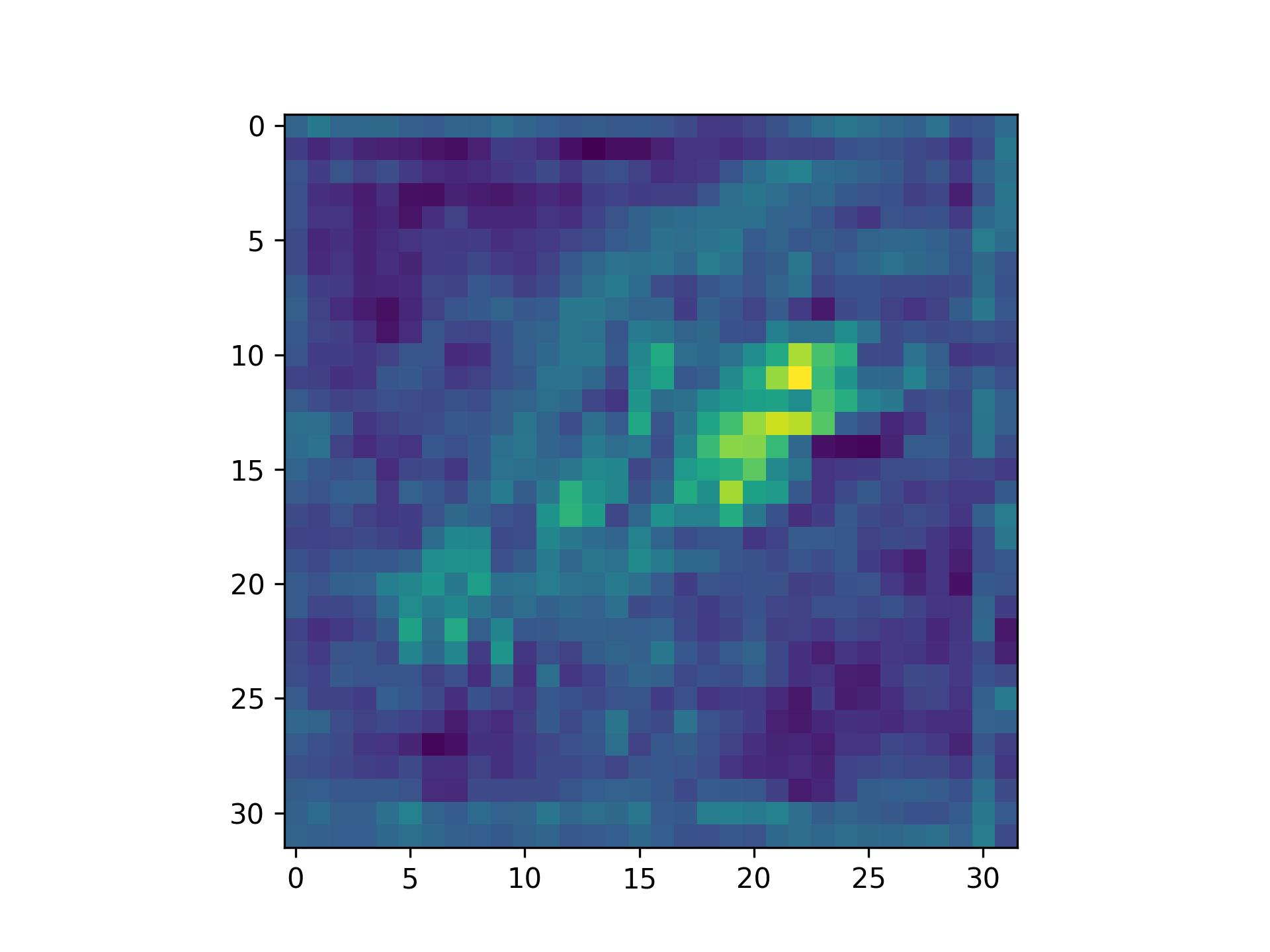}\\
  (a)
	\end{minipage}
	\begin{minipage}{0.24\linewidth}
		\centering
		\includegraphics[width=\linewidth]{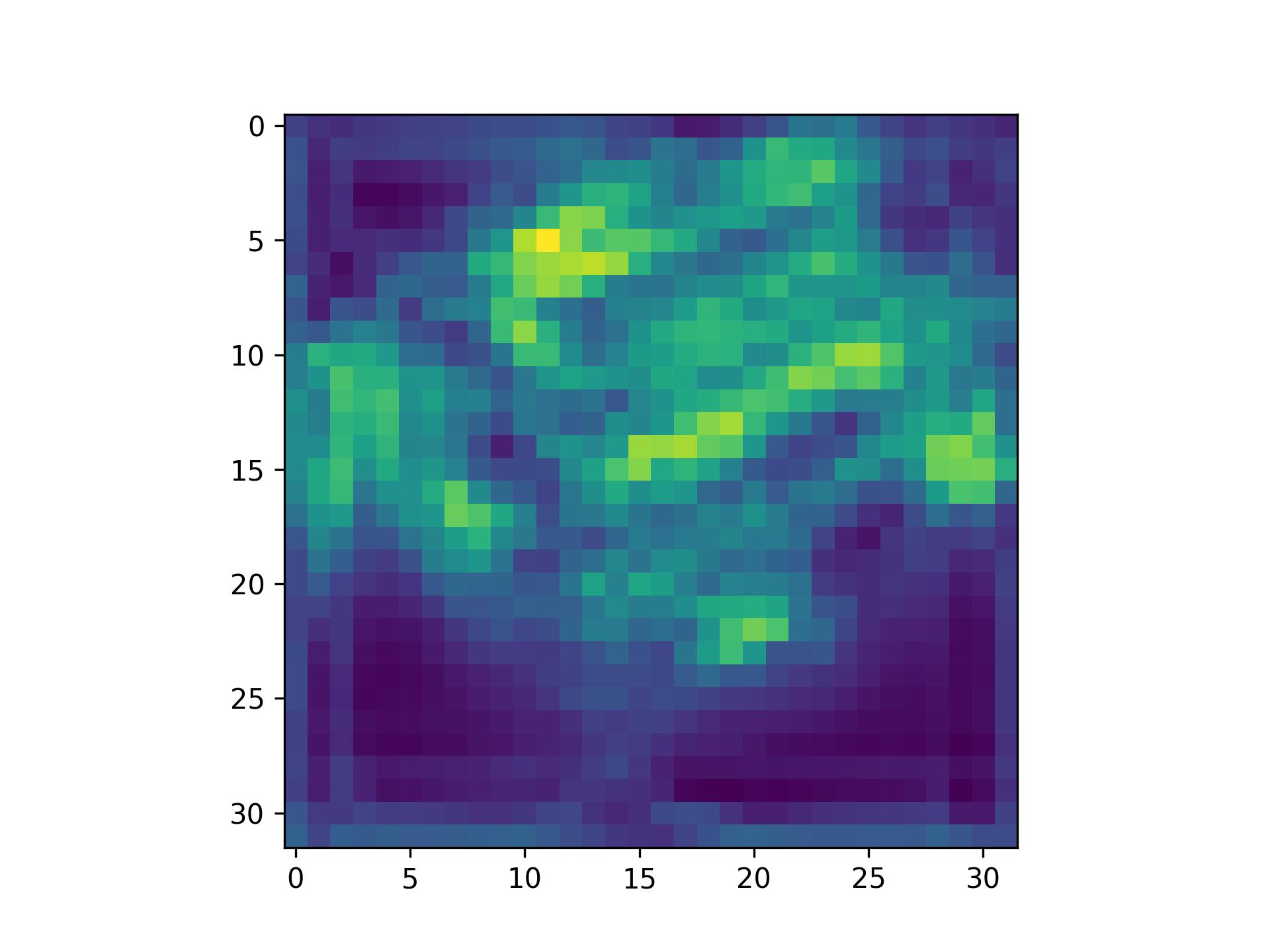}\\
  (b)
	\end{minipage}
 \begin{minipage}{0.24\linewidth}
		\centering
		\includegraphics[width=\linewidth]{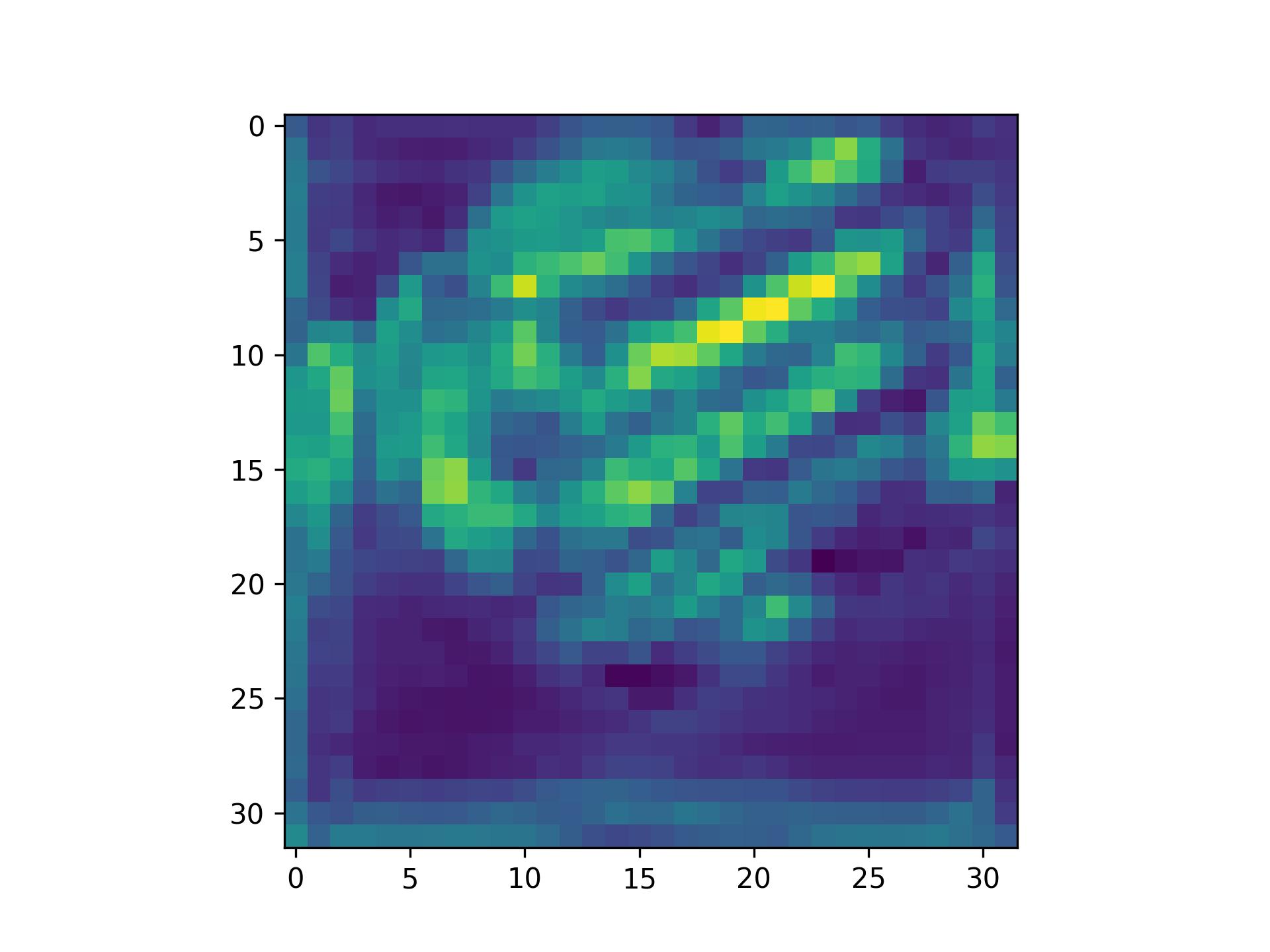}\\
  (c)
	\end{minipage}
 \begin{minipage}{0.24\linewidth}
		\centering
		\includegraphics[width=\linewidth]{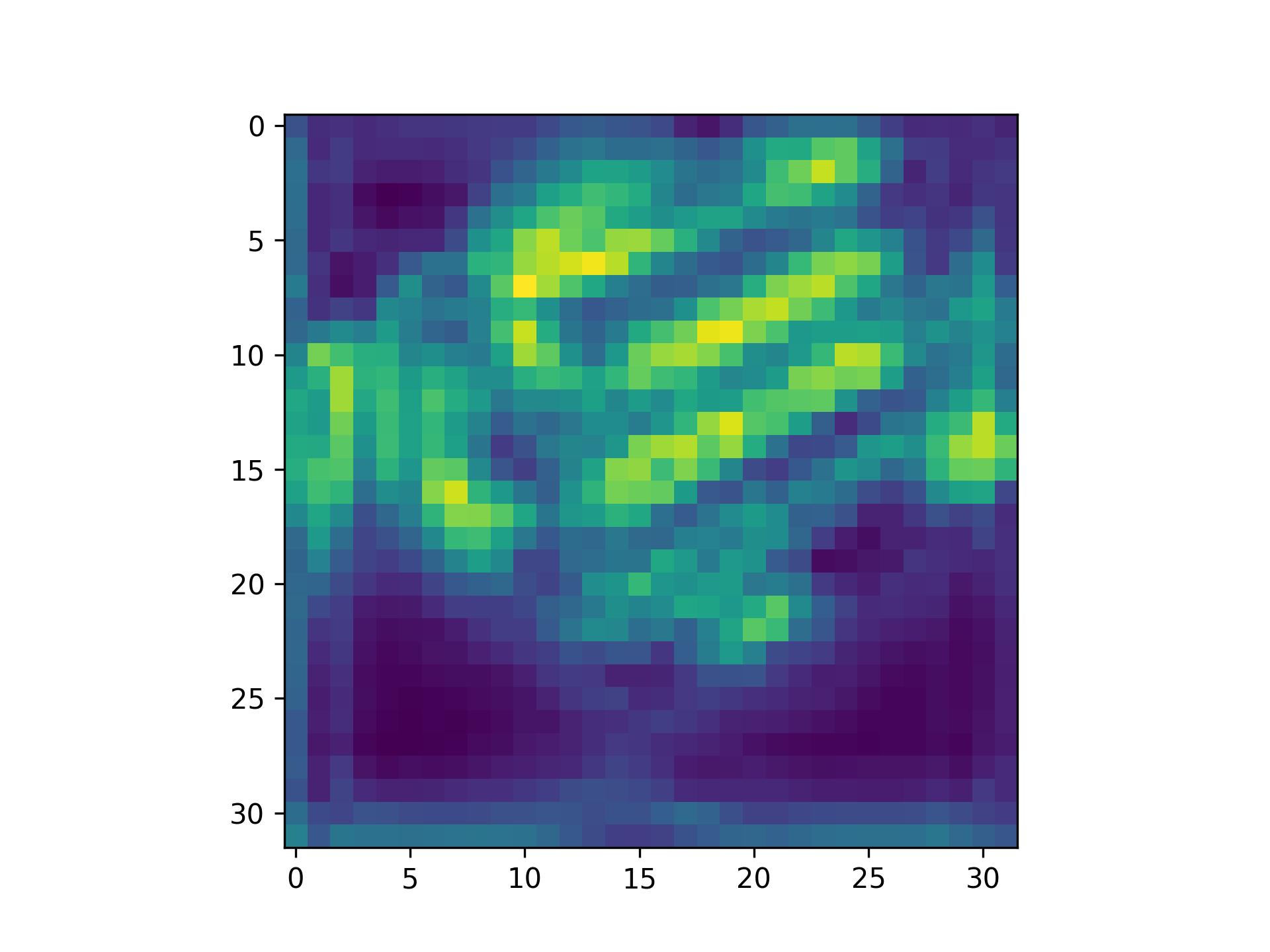}\\
  (d)
	\end{minipage}
 \caption{Feature maps comparison of (a) Original Method without MAN, (b) MAN with only EMA, (c) MAN with only learnable bias and (d) MAN with both EMA and learnable bias. The feature map are obtained using ResNet-32 (K=16) on CIFAR-10.}
 \label{Figure 5}
\end{figure}

As depicted in Fig. \ref{Figure 5}(a), we can observe that the original method without the addition of MAN learns very limited and chaotic features. After adding the EMA method alone, the features it learns are clearly more characteristic and defined, indicating that it indeed receives some global features from the information in the subsequent block. However, there are still many blurry features, which may be due to the EMA parameter update method \cite{34}, causing information imbalance due to feature differences between different local blocks. When we add learnable bias to the original method alone, the features it learns are more concentrated and clear, indicating that the learnable bias significantly enhances the learning ability of the current local block's hidden layer, but it still lacks some details because it does not receive more global information from the subsequent block. When we add the EMA method and learnable bias simultaneously, the learned features are not only clear and specific, but also more comprehensive in detail, proving that their learning abilities are complementary, and greatly improving the performance of the original method. The feature visualization in Fig. \ref{Figure 5} verifies our previous thoughts.

\subsection{The Effectiveness of EMA}

To verify the effectiveness of the EMA update method in MAN, we conduct further ablation experiments. As shown in Table \ref{Table 5}, using only the parameters of the next layer in MAN results in very limited improvement. In contrast, employing the EMA method to utilize the parameters of the next layer leads to more significant enhancements.
This is because the subsequent local block and the current local block learn different features and have different learning objectives, leading to minimal improvement. However, using the EMA method can gradually promote information exchange between local blocks and reduce parameter fluctuations during training, leading to smoother parameter updates. This beneficial interaction brings significant improvements to MAN.

\begin{table}[htbp]
	\centering
 \caption{Abalation study of EMA. (a) Using DGL as baseline and ResNet-32 (K=16) as backbone on the CIFAR-10 dataset. (b) Using InfoPro as baseline and ResNet-101 (K=4) as backbone on the ImageNet dataset. Parameter signifies whether each local block utilizes the parameters of the adjacent next layer connected to it.}
\begin{minipage}{0.49\linewidth}
\begin{tabular}{ccc} 
\hline Parameter & EMA & Test Error \\
\hline $\times$ & $\times$ & 14. 08 \\
$\checkmark$ & $\times$ & $\boldsymbol{12.93}$ \\
$\checkmark$ & $\checkmark$ & $\boldsymbol{11.07}$ \\
\hline
\multicolumn{3}{c}{(a)}
\end{tabular}
\end{minipage}
\begin{minipage}{0.49\linewidth}
\begin{tabular}{cccc} 
\hline Parameter & EMA & Top1-Error & Top5-Error \\
\hline $\times$ & $\times$ & 22.81 & 6.54 \\
$\checkmark$ & $\times$ & $\boldsymbol{22.64}$ & $\boldsymbol{6.41}$ \\
$\checkmark$ & $\checkmark$ & $\boldsymbol{22.09}$ & $\boldsymbol{6.07}$ \\ 
\hline
\multicolumn{4}{c}{(b)}
\end{tabular}
\end{minipage}
\label{Table 5}
\end{table}

\subsection{Linear Separability Analysis} 

\begin{figure}[H]
    \centering
    \includegraphics[width=0.9\textwidth]{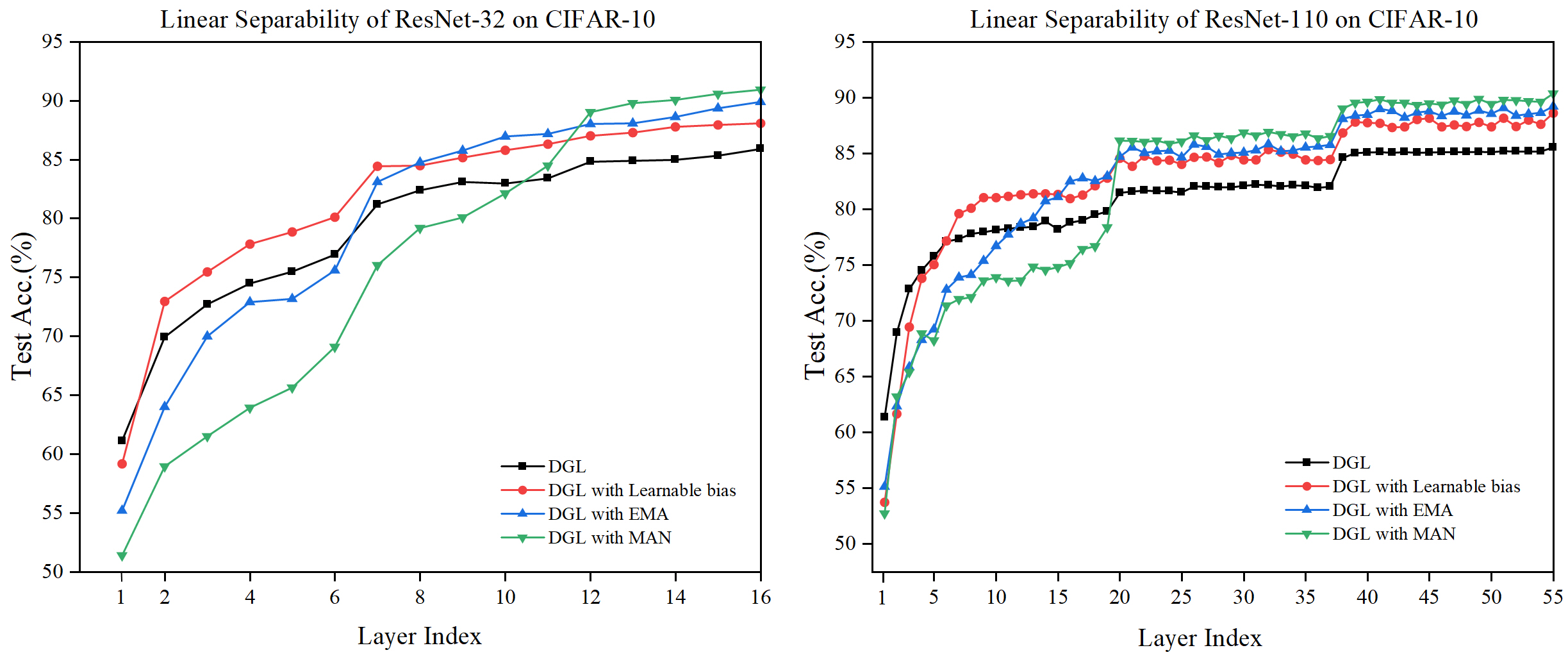}
    \caption{Comparison of layer-wise linear separability across different learning rules on ResNet-32 and ResNet-110.}
    \label{Figure 6}
\end{figure}

To demonstrate the effectiveness of our Momentum Auxiliary Network, we freeze the parameters of the main network and train a classifier for each local block to obtain the classification accuracy of each local block. As depicted in Fig. \ref{Figure 6}, we use DGL \cite{28} as a baseline, and after integrating our method, the accuracy of the earlier layers decrease, while the accuracy of the middle and later layers significantly improve. The decrease in accuracy of the earlier layers suggests that they have learned more generalized features. Although these features are not beneficial for optimizing local objectives, they are beneficial from a global perspective. The middle and later layers, receiving these globally beneficial features, have seen a significant increase in accuracy. These results illustrate that our proposed method has facilitated information interaction between gradient-isolated local blocks. This addresses the shortsightedness issue existing in the current supervised local learning field and significantly enhances their performance.

\subsection{Representation Similarity Analysis}

\begin{figure}[H]
    \centering
    \includegraphics[width=0.9\textwidth]{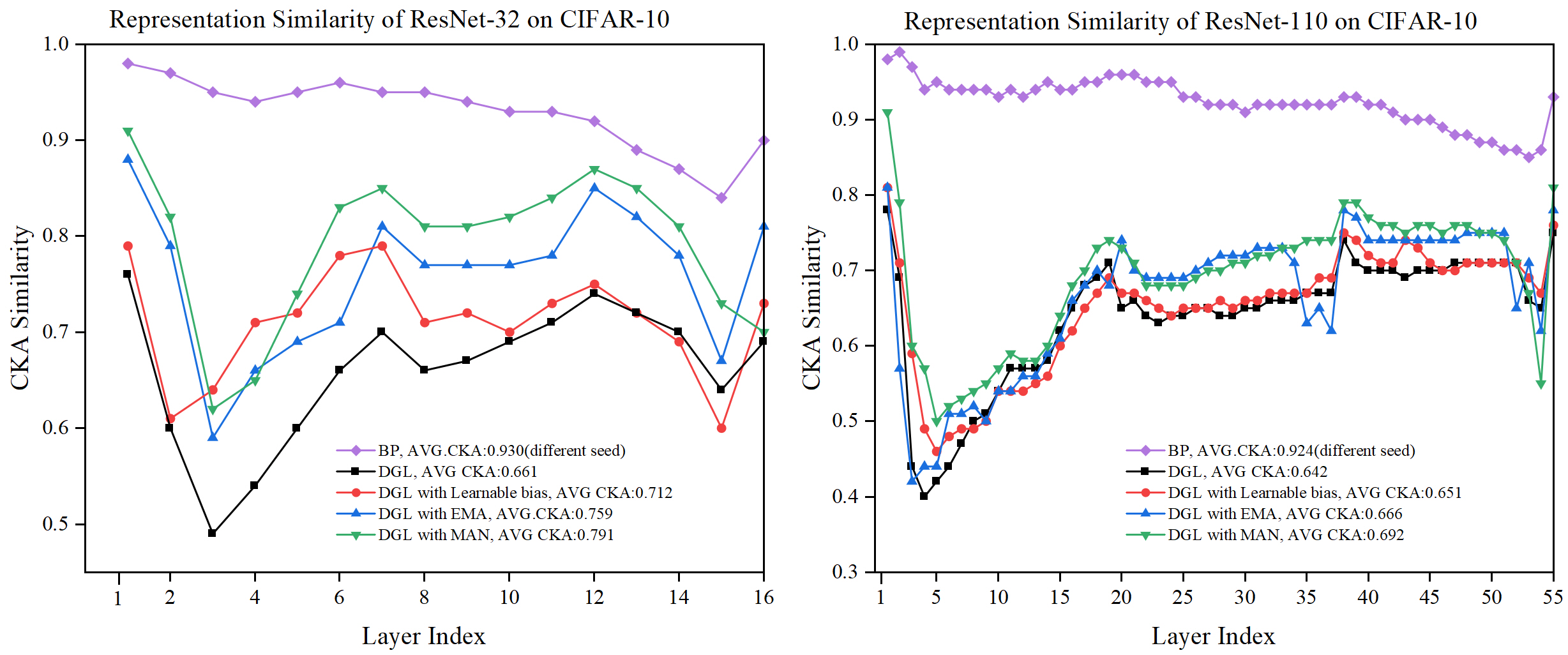}
    \caption{Assessment of Similarity in Layer-wise Representations. We use Centered Kernel Alignment (CKA) {\cite{kornblith2019similarity}} to quantify the degree of similarity in the layer-wise representations between the E2E backpropagation (BP) and our proposed MAN.}
    \label{Figure 7}
\end{figure}

To further validate the efficacy of MAN, we conduct a Centered Kernel Alignment (CKA) experiment \cite{kornblith2019similarity}. CKA serves as a metric to assess the similarity between feature representations. If a method's CKA score is closer to 1 in relation to the E2E training method, it indicates that the method's feature learning process is more aligned with that of E2E training. As depicted in Fig. \ref{Figure 7}, we use DGL \cite{28} as a baseline and incorporate our method. It can be observed that whether adding the EMA method or learnable bias alone, the CKA scores significantly improve compared to the original method. When both the EMA and learnable bias are used in MAN, the CKA score further improves and performs more stably. Notably, the most significant increase in the CKA score occurs in the early and late layers. In conjunction with our previous analysis of the linear separability experiment, this is because if the early layers' learning method is closer to E2E training, they will focus more on learning general features to optimize the global objective, rather than focusing narrowly on local objectives. While this may result in poorer classification capabilities in the early layers, it greatly contributes to the overall performance improvement of the network. Through the analysis of images and experimental results, it can be demonstrated that MAN enables information interaction between gradient-isolated local blocks, solving the myopia problem present in current supervised local learning methods.

\section{Conclusion}
This study addresses the performance disparities between traditional supervised local learning and end-to-end (E2E) methods in deep learning. We introduce a versatile Momentum Auxiliary Network to tackle the short-sighted problem in the early optimization work related to supervised local learning, facilitating information exchange between gradient-isolated local blocks. We integrate our Momentum Auxiliary Network into three advanced supervised local learning approaches and evaluate their performance across network architectures with varying depths on four widely adopted datasets. The results demonstrate our method's ability to significantly enhance the ultimate output performance of original supervised local learning methods. Particularly when combined with InfoPro \cite{27}, our method significantly reduces GPU memory usage while consistently maintaining performance levels closely aligned with E2E approaches.

\noindent \textbf{Limitations and future works:} Despite the superior performance on large-scale problems like ImageNet \cite{32}, our method still performs less accurately than E2E on some conventional image classification datasets. This may be due to the MAN using too few information interaction layers when the network is divided into a larger number of local blocks. In future work, we could explore deepening these information interaction layers to achieve better precision performance.

\section*{Acknowledgement}
This work was supported by the National Natural Science Foundation of China (52278154), the Natural Science Foundation of Jiangsu (BK20231429), the Fundamental Research Funds for the Central Universities (2242024RCB0008), and assupport from the program of Zhishan Young Scholar of Southeast University.

\bibliographystyle{splncs04}
\bibliography{11_references}


\end{document}


\title{\paperTitle}
\author{\authorBlock}
\maketitlesupplementary

\section{Appendix Section}
Supplementary material goes here.

{\small
\bibliographystyle{ieee_fullname}
\bibliography{11_references}
}